\documentclass[a4paper,conference]{ieeeconf}

\usepackage{cite}

\usepackage[pdftex]{graphicx}
\graphicspath{{gfx/}}
\DeclareGraphicsExtensions{.pdf,.jpeg,.jpg,.png}

\usepackage{tikz}

\usepackage{amsmath}
\usepackage{array}
\usepackage[hyphens]{url}

\usepackage{adjustbox}
\newcommand{\includegraphicsfillbox}[3]
{\bgroup
  \dimen1=#1\relax
  \dimen2=#2\relax
  \sbox0{\includegraphics[width=#1]{#3}}%
  \ifdim\ht0>\dimen2
    \dimen0=\dimexpr \ht0-\dimen2\relax
    \adjustbox{clip=true,trim=0pt 0.5\dimen0 0pt 0.5\dimen0}{\usebox0}%
  \else
    \sbox0{\includegraphics[height=#2]{#3}}%
    \ifdim\wd0>\dimen1
      \dimen0=\dimexpr \wd0-\dimen1\relax
      \adjustbox{clip=true,trim=0.5\dimen0 0pt 0.5\dimen0 0pt}{\usebox0}%
    \else
      \usebox0
    \fi
  \fi
\egroup}

\usepackage{tikz}
\usetikzlibrary{shapes,arrows}
\usetikzlibrary{calc}

\newcommand{\SymbRouter}{%
  \raisebox{-0.2\fontcharht\font`O}{%
    \resizebox{!}{1.3\fontcharht\font`O}{%
      \begin{tikzpicture}
        \draw [-stealth, thick] ($0.25*(1,0)$) -- (0.02,0);
        \draw [-stealth, thick] ($0.25*(-1,0)$) -- (-0.02,0);
        \draw [-stealth, thick] (0,0.02) -- ($0.25*(0,1)$);
        \draw [-stealth, thick] (0,-0.02) -- ($0.25*(0,-1)$);
        
        \draw (0,0) circle (0.3);
      \end{tikzpicture}%
    }%
  }%
}

\usepackage{hyperref}
\usepackage{cleveref}
\crefname{paragraph}{paragraph}{paragraphs}
\Crefname{paragraph}{Paragraph}{Paragraphs}

\usepackage{booktabs}
\usepackage{xcolor}
\usepackage{textcomp}

\usepackage{listings}

\makeatletter
\def\lst@makecaption{%
  \def\@captype{table}%
  \@makecaption
}
\makeatother

\AtBeginDocument{}

\crefname{lstlisting}{listing}{listings}
\Crefname{lstlisting}{Listing}{Listings}

\definecolor{delim}{RGB}{20,105,176}
\definecolor{numb}{RGB}{106, 109, 32}
\definecolor{string}{rgb}{0.64,0.08,0.08}

\lstdefinelanguage{json}{
    numbers=left,
    numberstyle=\small,
    frame=single,
    rulecolor=\color{black},
    showspaces=false,
    showtabs=false,
    breaklines=true,
    postbreak=\raisebox{0ex}[0ex][0ex]{\ensuremath{\color{gray}\hookrightarrow\space}},
    breakatwhitespace=true,
    basicstyle=\ttfamily\small,
    upquote=true,
    morestring=[b]",
    stringstyle=\color{string},
    literate=
     *{0}{{{\color{numb}0}}}{1}
      {1}{{{\color{numb}1}}}{1}
      {2}{{{\color{numb}2}}}{1}
      {3}{{{\color{numb}3}}}{1}
      {4}{{{\color{numb}4}}}{1}
      {5}{{{\color{numb}5}}}{1}
      {6}{{{\color{numb}6}}}{1}
      {7}{{{\color{numb}7}}}{1}
      {8}{{{\color{numb}8}}}{1}
      {9}{{{\color{numb}9}}}{1}
      {\{}{{{\color{delim}{\{}}}}{1}
      {\}}{{{\color{delim}{\}}}}}{1}
      {[}{{{\color{delim}{[}}}}{1}
      {]}{{{\color{delim}{]}}}}{1},
}

\newcommand\YAMLcolonstyle{\color{red}\mdseries}
\newcommand\YAMLkeystyle{\color{black}\bfseries}
\newcommand\YAMLvaluestyle{\color{blue}\mdseries}

\makeatletter

\newcommand\language@yaml{yaml}

\expandafter\expandafter\expandafter\lstdefinelanguage
\expandafter{\language@yaml}
{
  keywords={true,false,null,y,n},
  keywordstyle=\color{darkgray}\bfseries,
  basicstyle=\YAMLkeystyle,                                 
  sensitive=false,
  comment=[l]{\#},
  morecomment=[s]{/*}{*/},
  commentstyle=\color{purple}\ttfamily,
  stringstyle=\YAMLvaluestyle\ttfamily,
  moredelim=[l][\color{orange}]{\&},
  moredelim=[l][\color{magenta}]{*},
  moredelim=**[il][\YAMLcolonstyle{:}\YAMLvaluestyle]{:},   
  morestring=[b]',
  morestring=[b]",
  literate =    {---}{{\ProcessThreeDashes}}3
                {>}{{\textcolor{red}\textgreater}}1     
                {|}{{\textcolor{red}\textbar}}1 
                {\ -\ }{{\mdseries\ -\ }}3,
}

\lst@AddToHook{EveryLine}{\ifx\lst@language\language@yaml\YAMLkeystyle\fi}
\makeatother

\newcommand\ProcessThreeDashes{\llap{\color{cyan}\mdseries-{-}-}}

\usepackage[disable]{todonotes}

\usepackage{flushend}

\usepackage{background}

\newcommand\copyrighttext{%
  \footnotesize \textcopyright 2022 IEEE. Personal use of this material is %
  permitted. Permission from IEEE must be obtained for all other uses, in any %
  current or future media, including reprinting/republishing this material for %
  advertising or promotional purposes, creating new collective works, for resale %
  or redistribution to servers or lists, or reuse of any copyrighted component %
  of this work in other works. %
  DOI:%
  \href{https://doi.org/10.1109/SII52469.2022.9708884}{10.1109/SII52469.2022.9708884}%
}

\newcommand\copyrightnotice{%
  \begin{tikzpicture}[remember picture,overlay]%
    \node[anchor=north,yshift=-24pt] at (current page.north) {\fbox{\parbox{\dimexpr\textwidth-\fboxsep-\fboxrule\relax}{\copyrighttext}}};%
  \end{tikzpicture}%
}

\newcommand\makecopyrightnotice{%
  \backgroundsetup{opacity=1, scale=1, angle=0, color=black, contents={\copyrightnotice}}%
}

\IEEEoverridecommandlockouts                              
\title{\LARGE \bf
Proxying ROS communications --- enabling containerized ROS deployments in distributed multi-host environments
}

\author{Arne Wendt$^{1}$ and Prof. Dr.-Ing. Thorsten Schüppstuhl$^{2}$
\thanks{$^{*}$All authors are with the Institute for Aircraft-Production-Technology,
        Hamburg University of Technology, 21073 Hamburg, Germany}%
\thanks{$^{1}${\tt\small 0000-0002-5782-3468 \textnormal{\small--} arne.wendt@tuhh.de}}%
\thanks{$^{2}${\tt\small 0000-0002-9616-3976 \textnormal{\small--} schueppstuhl@tuhh.de}}%
}

\begin{document}
\maketitle

\makecopyrightnotice

\thispagestyle{empty}
\pagestyle{empty}

\begin{abstract}
  With the ability to use containers at the edge, they pose a unified solution
  to combat the complexity of distributed multi-host ROS deployments, as well as
  individual ROS-node and dependency deployment. The bidirectional communication
  in ROS poses a challenge to using containerized ROS deployments alongside
  non-containerized ones spread over multiple machines though. We will analyze
  the communication protocol employed by ROS, and the suitability of different
  container networking modes and their implications on ROS deployments. Finally,
  we will present a layer 7 transparent proxy server architecture for ROS, as a
  solution to the identified problems. Enabling the use of ROS not only in
  containerized environments, but proxying ROS between network segments in
  general.
\end{abstract}


%

\section{Introduction}\label{intro}

Software containers as a technology gained traction mainly as a solution to
problems encountered in cloud computing.\cite{TheHistoryof2015} Providing
isolation between processes and their environments, they enable lightweight
multi-tenancy and efficient resource sharing.\cite{ContainersandCloud2014}
Current container technology does offer more though (compare e.g.
\cite{Docker2015}): Applications with all their dependencies (services,
configuration, etc.) can be bundled as a single artifact, seriously simplifying
deployment processes. Shipping all dependencies bundled within one artifact,
further allows for greatly simplified (even immutable) infrastructure -- a host
with just a container runtime -- reducing host-deployment and -maintenance
efforts. With intercompatible container runtimes being available for all major
operating systems (OSs) (Linux, Windows, OSX), a containerized application can be
deployed virtually independent of the host machine and OS.

Apart from the increased security through process- and usually additional
network-isolation from the host machine, especially the benefits in deployment
and cross-platform compatibility have made containers (and \emph{docker}
especially) a tool for reproducible deployments not only in cloud computing, but
research in general, as proposed in \cite{Anintroductionto2015}, and even
robotics research in particular, as proposed in \cite{TrytoStart2019}. As
identified in \cite{ROSandDocker2017}, a major advantage of bundling
dependencies in robotics development lies in consistently reproducible
deployments of even cutting edge algorithms and software modules, not available
as stable distributions through package management systems.

ROS as a framework enables the composition of robotics applications as
distributed systems on heterogenous hardware. While this trait gives a high
flexibility in development and enables optimal hardware choice and resource
utilization, multi-host distributed systems require serious maintenance efforts.
To mitigate compatibility issues, ROS provides a package management solution as
specially maintained and versioned repositories, and automated dependency
fetching using \emph{rosdep}\cite{rodepROSWiki} to support the deployment
process. \emph{roslaunch}, on the other hand, is ROS' own solution to combating
runtime complexity, 
supporting node spawning, parametrization and supervision, as well as setting
parameters and supporting remapping \cite{roslaunchROSWiki}. While all the latter simplifies the
deployment of ROS nodes, still the ROS distribution itself and
additional dependencies like services and necessary files have to be maintained
manually. As shown in \cite{VirtualizationonInternet2017}, containers are a
technology suitable to be used on resource constrained systems with negligible
performance penalties. With this ability to use containers at the edge, they pose
a unified solution to combat the complexity of distributed ROS-system deployment
over different hardware, as well as ROS-node an dependency deployment with
cutting edge technology implementations not available through package
management.

The bidirectional communication and establishment of communication channels
required by ROS poses a challenge to using ROS deployments spread over multiple
machines, mixing containerized and non-containerized nodes. We will first analyze
the suitability of different container networking modes for ROS communication,
as well as their possible implications on the final deployment. Using the
obtained information we will derive the concrete problem of using ROS in
containers in true distributed multi-host environments, and develop a suitable
solution.


\emph{For the remaining work we will focus on \emph{docker} as container
runtime, as it is available for all major hardware platforms and operating
systems, optimally reflecting the idea of heterogenous host systems. This choice
influences the following analysis in regard to the provided networking modes and
their specific implementation; most runtimes do provide equivalent networking
modes though (compare e.g. \cite{NetworkingoverviewDocker,UseIPvlannetworks}
with \cite{NetworksLXD} and \cite{systemd.nspawn5}). The proposed solution will be
runtime agnostic.}

\section{ROS and docker -- related works}\label{relatedworks} Official ROS
\emph{docker} images are published to \emph{docker hub} since the release of ROS
\emph{Jade}\cite{ROSandDocker2017}, introduced at ROScon
2015\cite{ROS+Docker2015}. The ROS Wiki officially introduces docker in
\cite{dockerTutorialsDocker}, but covers the use of container on a single host
only. The complementing documentation for ROS and docker networking at
\cite{dockerTutorialsNetwork} is not existent. While
\cite{RobotOperatingSystem2017} devotes a whole chapter\cite{ROSandDocker2017}
to docker, networking outside a single host is not covered, though the author
proposes \emph{docker swarm} networking for connecting multiple hosts at ROSCon
2015\cite{ROS+Docker2015}.

While acknowledged as a tool for establishing reproducibility in robotics
applications, we can mostly find publications on benefits of using \emph{docker}
with ROS in academic use cases and descriptions of uses in education like
\cite{TrytoStart2019,ROSLabSharingROS2019}. While we can find
\cite{AmodularCPS2017} as a solitary example, publicly available works using
\emph{docker} with ROS for \emph{distributed deployments} seem sparse. None of
the above examples exhibits and covers the challenges of building distributed
ROS deployments with containerized nodes. As we will discover in
\cref{dockernwros}, the challenge in containerizing ROS nodes is a
communications problem. We will thus briefly consider works combining cloud
computing with ROS, as the encountered problems from network partitioning are
assumed to be similar. While in \cite{CloudroboticsA72016} an OpenStack-based
cloud is used for data processing and storage, this cloud is a local deployment
on the same network as the remaining nodes, thus not incurring any problems of
traditional cloud computing. We can find an actual cloud computing setup in
\cite{CloudRoboticswith2020}, overcoming network boundaries by using
\emph{rosbridge} an a custom \emph{cloud bridge}. Apart from \emph{rosbridge} --
identified as the most common tool to enable cloud robotics applications with
ROS by \cite{Therisingprospects072013} -- we can find
\emph{ROSLink}\cite{ROSLinkBridgingROS2017} as another tool to solve the problem
of connecting ROS applications over multiple network segments. All these
solutions do not behave transparently for a ROS deployment, but have to be
interfaced explicitly with custom, application specific protocols.


\section{Docker networking and ROS}\label{dockernwros}
To understand the problems and limitations when using ROS in containers, we will
first introduce the operating principle of the ROS middleware. With this
foundation we will evaluate the fitness for containerized ROS applications of
the currently available networking modes provided by \emph{docker}.

\subsection{ROS communications}\label{roscomm} The ROS middleware is composed of
two different parts: An \emph{Extensible Markup Language Remote Procedure Call
(XMLRPC)}-based management protocol and the actual data transmission protocol
\emph{TCPROS}. Following we will analyze both protocols separately.

\begin{figure}[!t]
  \centering
  \vspace{1\baselineskip}
  \includegraphics[width=0.85\linewidth]{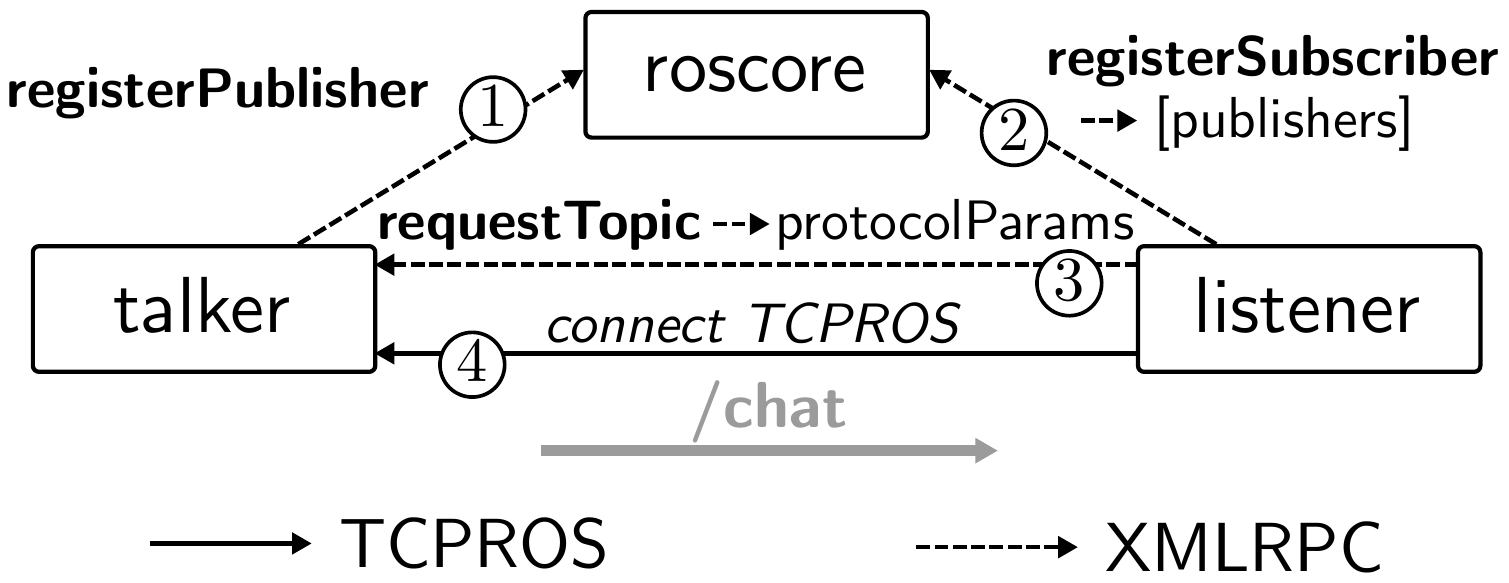}
  \caption{Order of communications in ROS: Advertising topic \emph{/chat} by
  node \emph{talker} and subscribing from node \emph{listener}. Arrows showing
  direction of communications channel establishment, and XMLRPC method call and
  return value.}
  \label{roscomm:img:order}
\end{figure}

\subsubsection{XMLRPC -- Master- \& Slave-API}
ROS defines two XMLRPC-based Application Programming Interfaces (APIs) for
management purposes; a \emph{Master-}\cite{ROSMaster_APIROS} and
\emph{Slave-API}\cite{ROSSlave_APIROS}. XMLRPC uses the \emph{Hypertext Transfer
Protocol (HTTP)} as underlying transport.

\paragraph{Master-API}
The \emph{Master-API} is implemented by the ROS master. Its address has to be
known to all nodes participating in a ROS network. Nodes consume this API to
(un-) register topic publications and subscriptions, services, as well as
performing topic-, service- and node-information-lookup. When registering, nodes
report their slave API endpoint to the master, using an explicitly configured
hostname or IP-address. A connection to consume the \emph{Master-API} is
initiated by a node.

\Cref{roscomm:img:order} shows invocation of the \emph{Master-API} for the
\emph{talker} node, registering as publisher for the \emph{/chat} topic. As step
2, \emph{listener} calls the \texttt{registerSubscriber} method on the
\emph{Master-API} to register itself as a subscriber to the \emph{/chat} topic,
and retrieve a list of \emph{Slave-API}-XMLRPC-endpoint-addresses of nodes
publishing the topic.

\paragraph{Slave-API}
Each node provides a \emph{Slave-API}. This API is consumed by the master as
well as other nodes. Apart from querying runtime information about a node, the
master consumes this API for management purposes like updating parameters and
shutting down nodes, as well as to notify nodes about new publishers of their
subscriptions. Nodes mutually consume their \emph{Slave-APIs} to negotiate a
communication channel for topic-data transmission; the subscribing node
initiates the connection. A random TCP-port is allocated for the underlying HTTP
server.

\Cref{roscomm:img:order} shows the \emph{Slave-API} call for the
\texttt{requestTopic} method from \emph{listener} to \emph{talker}, yielding a
description (\texttt{protocolParams}) of a mutually supported communications
channel\footnote{See \cref{TCPROS} for limitations}.

\subsubsection{TCPROS}\label{TCPROS} While ROS supports the negotiation of the
protocol and implementation to use for data exchange, TCPROS is the only
officially provided and supported protocol. TCPROS supports data exchange for
topics and service calls; representing the last step and actual data exchange
(4) in \cref{roscomm:img:order}. Per node usually one TCPROS endpoint is
allocated on a random TCP-port, the connection is established by a subscribing
oder service-calling node. In contrast to the XMLRPC/HTTP based management
protocol, TCPROS does not include any routing information.\cite{ROSTCPROSROS}

\subsection{Docker networking-mode fitness for ROS}\label{dockernw} We will
cover the networking modes provided by a stock \emph{docker} installation. While
\emph{libnetwork} -- as provider and implementation of virtual networking in
\emph{docker} -- does support plugins to extend its functionality, their
deployment on host machines adds additional efforts, largely invaliding the
motivation of containerizing applications. The available network modes are
documented in brief at \cite{NetworkingoverviewDocker}; while omitting
\emph{ipvlan} for unknown reasons, its documentation is available at
\cite{UseIPvlannetworks}. Following, the need for bidirectional communication
establishment as seen in \cref{roscomm}, will be the main focus of our
discussion. Discussed networking modes will be \emph{bridged}, \emph{host},
\emph{ipvlan} and \emph{macvlan}, as well as \emph{overlay} networking.
\emph{ipvlan} and \emph{macvlan} target the use case of connecting containers to
external (outside the host machine) networks. For these networking modes we will
therefore as well discuss the stock IP-address management (IPAM) mechanism and
behavior, as these impact the integration mechanisms of containers in external
networks.

\subsubsection{Bridged}\label{dockernw:bridged} 
\begin{figure}[!t]
  \centering
  \vspace{0.5\baselineskip}
  \includegraphics[width=0.9\linewidth]{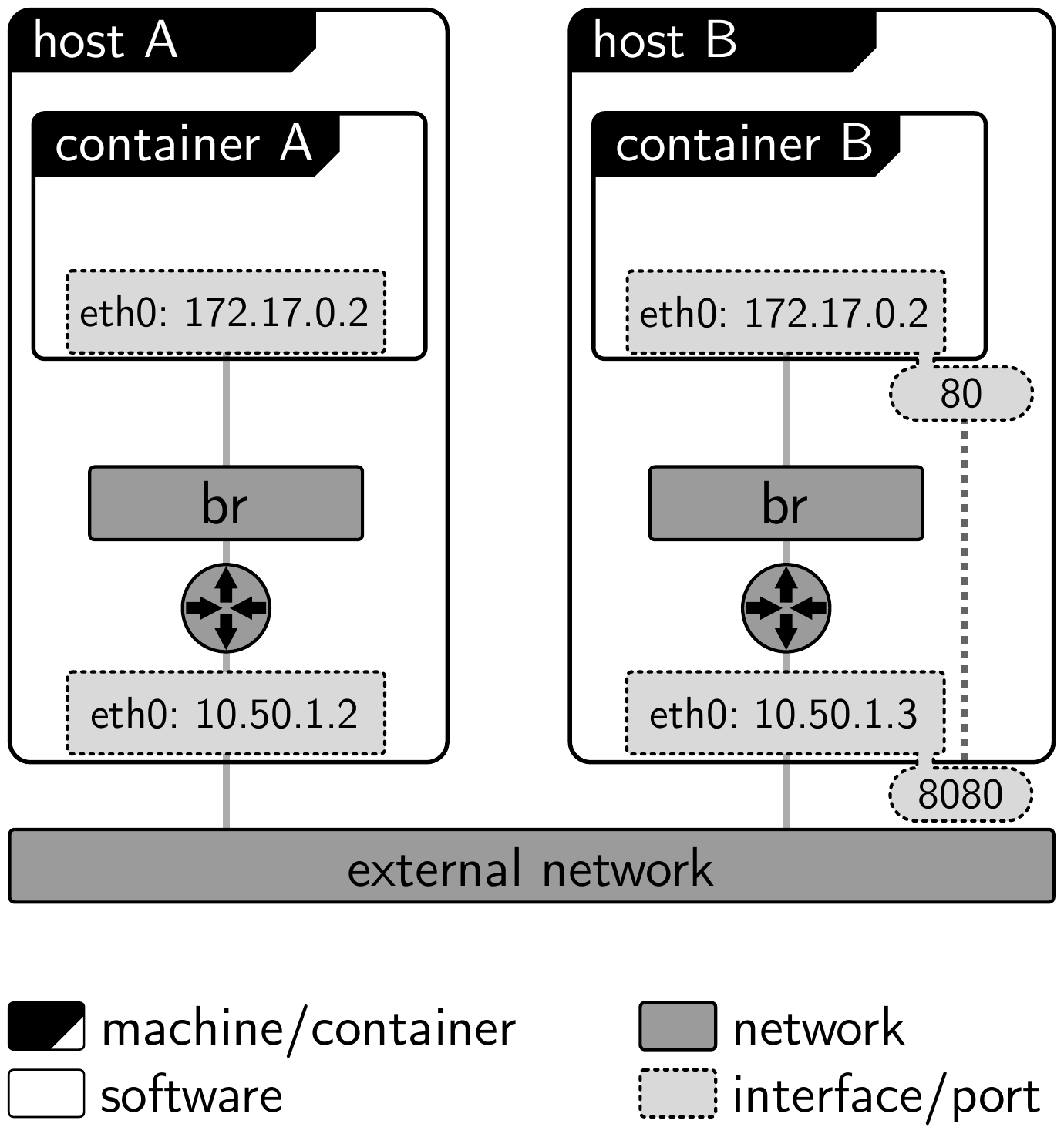}
  \caption{\emph{docker} default bridge networking topology}
  \label{container:img:br}
\end{figure}

\emph{Bridged} is the default
network mode used by \emph{docker}. The operation principle is shown in
\cref{container:img:br}. A network bridge \emph{br} is created on the host
machine \emph{host A/B}. The virtual network interfaces \emph{eth0} of
containers \emph{container A/B} connect to the bridge acting as a network
switch. Connectivity from the containers to an \emph{external network} segment
is provided by a router \SymbRouter, performing Network Address Translation
(NAT) between \emph{br} a host network interface (e.g. \emph{eht0}). Each host
can communicate with all containers running on that host, e.g. \emph{host A}
with \emph{container A}, and \emph{host B} with \emph{container B}. As the host
performs NAT routing, containers can reach all network targets on \emph{external
network} the host itself can reach, e.g. \emph{container A} can reach \emph{host
A} and \emph{host B}. Containers, being located behind a router, are not
directly reachable from the \emph{external network}, e.g. \emph{host A} and
\emph{container A} cannot reach \emph{container B}. Exposing applications from
within a container to the \emph{external network} is performed by forwarding
ports from the host to the container. This technique is shown for \emph{host B}
and \emph{container B}: Forwarding port 80 from the container to port 8080 on
the host machine, allows participants on the \emph{external network} to access a
service on port 80 in \emph{container B} by connecting to port 8080 on
\emph{host B}.

ROS requires bidirectional connection establishment, which is not directly
satisfiable using \emph{bridge} networking. As described in \cref{roscomm}, the
ports allocated for communication by ROS are chosen at random. This behavior
prohibits the use of port forwarding to make containerized applications
available to an outside network segment, as the ports to forward are not
deterministic and not know a priori. Thus, using \emph{bridged} networking, it is
not directly possible to connect containerized ROS nodes to nodes on an external network.

\subsubsection{Host}
Using \emph{host} networking mode does not allocate virtual network interfaces
within a container, but grants direct access to the host network interfaces from
within a container.

Tough it does technically allow ROS nodes to be ran inside a container and
communicate with nodes on an external network, we evaluate \emph{host}
networking to not qualify as a universal solution for the following reasons: It
removes network isolation and thus a major portion of the benefits of
containerizing applications, and it is no portable solution as it is only
available on Linux based system \cite{Usehostnetworking}.

\subsubsection{ipvlan \& macvlan}
\emph{Ipvlan layer3-mode} uses a similar setup to \emph{bridge} networking, and
exhibits similar behavior with the major difference, that reverse routing is
technically possible when using static routes from the external network. Due to
this similarity we will not consider \emph{ipvlan layer3-mode} any further and
scope the following analysis to \emph{ipvlan layer2-mode}. \emph{Ipvlan} and
\emph{macvlan} networking modes each allocate a sub-interface on an interface of
the host machine, mapping this sub-interface into a container. While
\emph{macvlan} allocates an additional MAC-address on the interface,
\emph{ipvlan} does only allocate an additional IP-address. In each case, a
container gets direct access to the network of the underlying host/parent
interface, while retaining host and container network isolation and allowing
members of the external network to reach containers as if they were physical
hosts.

While both networking modes -- \emph{ipvlan layer3-mode} and \emph{macvlan} --
do in theory provide an optimal solution in terms of enabling direct network
access for containers, they do come with a caveat though: IP-address management.

\paragraph{IPAM}\label{ipam} Networking in \emph{docker} is implemented using
\emph{libnetwork}. Each networking mode is implemented by a
driver\cite{libnetworkDesign}. A driver is responsible for all aspects of
network-operation, including IP-address management (IPAM). IPAM in turn is
implemented by IPAM-drivers, owned and controlled by an instance of a network
driver\cite{libnetworkDesign,libnetworkIPAMDriver}. The available default
IPAM-drivers shipped with \emph{docker} do assign addresses from either a random
or user-defined subnet, but do not support address assignment using Dynamic Host
Configuration Protocol (DHCP).
Using \emph{macvlan/ipvlan} networking on multiple hosts on the same network
requires synchronization of subnets to guarantee connectivity, and
synchronization of disjoint address-pools to avoid address conflicts, over all
hosts; adding management efforts and an additional problem of configuration
distribution among hosts.

While a third party network driver with DHCP support is available at
\cite{dockernetdhcp}, and an experimental DHCP-IPAM-driver is available at
\cite{nerdalertlibnetwork,ExperimentalDockerLibnetwork}, we could not confirm
them to be working on all OSs targeted by \emph{docker}. As stated in
\cref{dockernw} we do not consider plugins to be a viable universal solution
(see \cref{proxy} for further information). Based on the behavior of the stock
IPAM-drivers we evaluate \emph{macvlan} and \emph{ipvlan} networks as not
suitable for containerized ROS applications. We believe address management to be
an issue to be tackled by either using DHCP, or while deploying the host
machines and their operating systems, and not to be solved by manual
configuration distribution at container instantiation time.

\subsubsection{Overlay}
Allows to build networks spanning multiple \emph{docker} instances on multiple
hosts, transparently connecting containers over host boundaries. While solving
the problem of connecting containerized ROS nodes on multiple hosts,
\emph{overlay} networking does not allow to communicate with non-containerized
nodes on different hosts and/or networks.

\subsection{Problem Statement}\label{problem}
Focusing on \emph{docker}, as the container runtime available for the most
common operating systems, from our analysis in \cref{dockernw} we can conclude,
that no stock-available networking mode is suitable to run containerized
ROS nodes, with our basic requirements:
\begin{enumerate}
  \item nodes distributed among different machines
  \item mix of nodes ran directly on host OS and containerized
  \item no additional configuration distribution and management
\end{enumerate}

While \emph{overlay} networking allows to satisfy requirement 1, it fails to
satisfy requirement 2 and 3. \emph{Macvlan} and \emph{ipvlan} networking can satisfy
requirements 1 and 2, while failing to satisfy requirement 3 with the provided
\emph{IPAM} implementation (see \cref{ipam} for details).

Consequently, we need to find a way to enable the use of ROS in containers,
while being able to communicate between containerized nodes and nodes ran
directly on the host OS, and removing additional \emph{out-of-band}
configuration and management. As \emph{out-of-band} we want to classify all
configuration an management that is not local to ROS and \emph{docker}, or not
supported by their distribution/deployment mechanisms; e.g. installing
third-party plugins would be considered \emph{out-of-band}, while pulling
container images from a registry would not (see e.g. \cref{ipam}).

\section{L7 ROS proxy}\label{proxy}

\begin{figure}[!t]
  \centering
  \vspace{0.5\baselineskip}
  \includegraphics[width=0.9\linewidth]{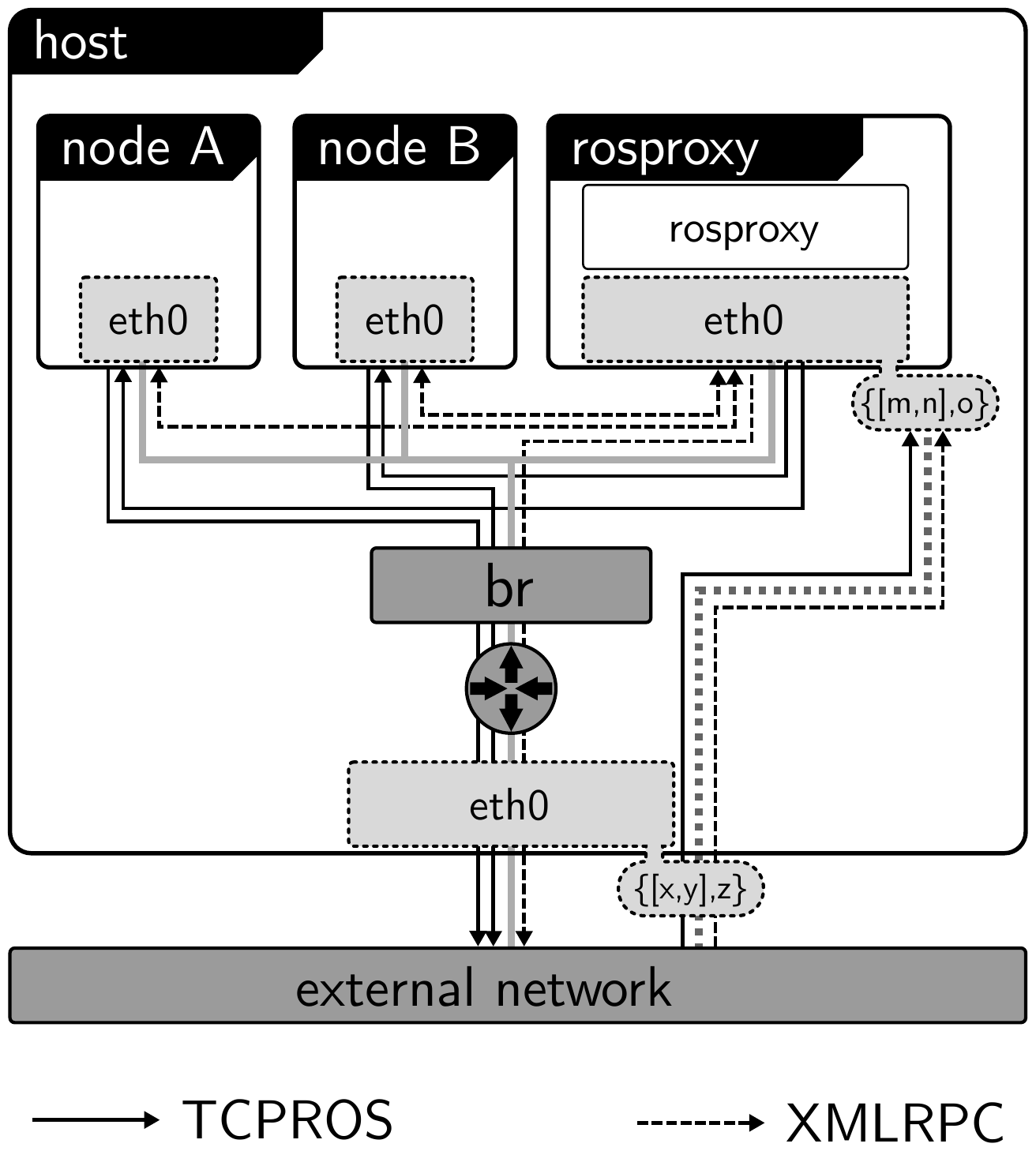}
  \caption{Networking topology and connections, when using containerized ROS
  nodes with proposed \emph{rosproxy}}
  \label{container:img:rosproxy}
\end{figure}

\begin{figure*}[!t]
  \centering
  \vspace{1\baselineskip}
  \includegraphics[width=0.85\linewidth]{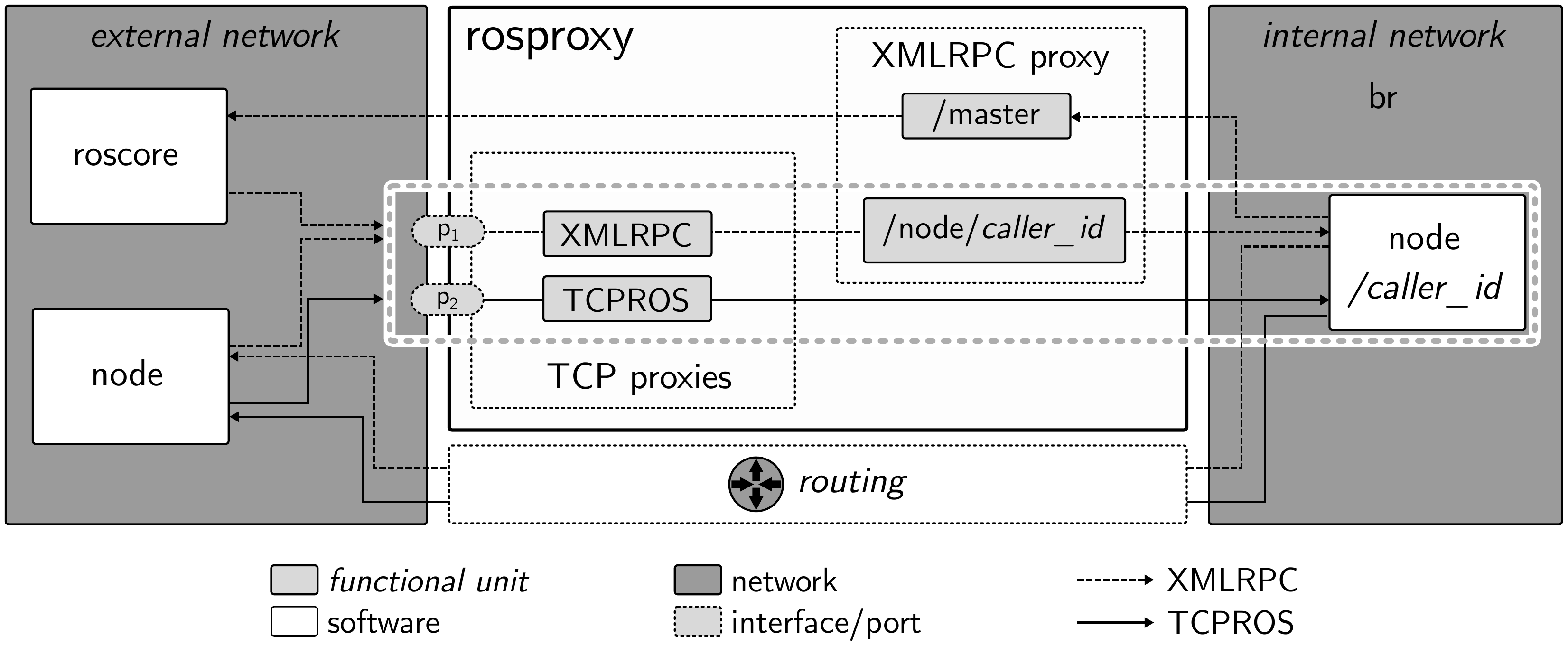}
  \caption{\emph{rosproxy} schematic operation principle and connection topology}
  \label{rosproxy:img}
\end{figure*}

As the container runtime cannot provide a suitable solution by itself, we have
to find a solution \emph{within} the bounds of ROS and \emph{docker}, on a
higher networking layer.

With \emph{bridged} networking and port forwarding, \emph{docker} provides a
networking mode with the ability to expose containerized applications on the
network, with no configuration to be shared and managed among multiple hosts. As
described in \cref{dockernw:bridged}, the main problem of using \emph{bridged}
networking is the non-deterministic allocation of ports used for ROS
communications. Finding a way to allocate all ports used for ROS communications
from a deterministic port range, would provide a foundation to use
\emph{bridged} networking with containerized ROS nodes. Allocating and
forwarding ports to the host machine would provide an additional benefit: Name-
and address-resolution will be scoped to the network of host machines and not
require routing rules and public (scoped to the network) name resolution
services for containers. We will further require our solution to be
\textbf{transparent} (i.e. no changes to node implementations and network
configuration of containers) for all nodes; containerized or not.



\subsection{Concept}
As a solution we propose a proxy\footnote{We will later find, that a proxy
server provides additional benefits outside of containerization.} for ROS
communications, on the application layer (layer 7 / L7), ran in a separate
container, with the capability of allocating all ports for ROS communication
from a specified range to be forwarded to the host. The concept is shown in
\cref{container:img:rosproxy}. A proxy server \emph{rosproxy} is ran in a
container, sharing a bridged network with containers running ROS nodes. A set
$\{ [m,n],o \}$ of a port $o$ and a range of ports $[m,n]$ from the proxy
container will be forwarded to the hosts interface \emph{eth0} as ports
$\{[x,y],z \}$. While containerized ROS nodes can connect directly to other
nodes on the network using TCPROS,
the \emph{rosproxy} shall allow the reverse and proxy TCPROS communication from
ports within the range $[m,n]$ to the relevant containerized nodes in containers
\emph{node A/B}. As seen in \cref{roscomm}, the latter is required for all
XMLRPC communications as well.

\subsubsection{Proxying XMLRPC}
Nodes report their XMLRPC-management-API to the ROS master, consisting of a
randomly allocated TCP-port and either a hostname or IP-address (explicitly
configured from environment variables \texttt{ROS\_HOSTNAME} or
\texttt{ROS\_IP}). To expose these API endpoints using the proxy, communication
from nodes to the master will have to be intercepted (cmp.
\cref{container:img:rosproxy}), and the reported XMLRPC endpoints rewritten to
the address of an endpoint provided by the proxy and reachable on the external
network. Requiring interception and modification of nodes XMLRPC calls to the
master, we can find a simple concept of \textbf{transparently} injecting the
proxy server into the communications: The proxy server will be set as the ROS
master address for all containerized nodes on a machine. On startup, nodes will
register with the master. Configuring the proxy as the address of the ros
master, we can intercept this communication. XMLRPC leveraging HTTP as transport
protocol, allows to dynamically allocate an HTTP endpoint on the proxy per node,
rewrite the nodes XMLRPC-API-address to this new endpoint, and proxy calls to
this endpoint back to the respective node. As documented in
\cite{ROSMaster_APIROS}, the calls reporting a nodes XMLRPC address (as
\texttt{caller\_api}), thus requiring interception and modification, are
\texttt{registerService}, \texttt{registerSubscriber},
\texttt{unregisterSubscriber}, \texttt{registerPublisher} and
\texttt{unregisterPublisher}. \cref{container:img:rosproxy} shows this behavior
of relaying all XMLRPC requests from nodes through the \emph{rosproxy}, as well
as relaying all calls to the nodes slave APIs through the proxy. All ingress
from the external network is over ports forwarded to the hosts network
interface.

\subsubsection{Proxying TCPROS}
TCPROS itself does not contain any connection information and does not need any
interception and data modification. To proxy TCPROS connections, a simple TCP
proxy, forwarding traffic from a port on the host to the port allocated for
TCPROS communications by a ROS node is sufficient. The TCPROS endpoint, as
allocated by a node, is reported either on service registration
(\texttt{service\_api} when calling \texttt{registerService} on the master API
\cite{ROSMaster_APIROS}) or as a protocol parameter in response to a
\texttt{requestTopic} call to a nodes slave API \cite{ROSSlave_APIROS}. To
effectively proxy TCPROS connections, these XMLRPC calls have to be intercepted
as well; instantiating a new TCP proxy for each TCPROS port reported by a node,
and rewriting the \texttt{service\_api} or relevant \texttt{ProtocolParam}
values in the XMLRPC call or response. As shown in
\cref{container:img:rosproxy}, outbound TCPROS connections from within a
container are routed directly by \emph{docker}, while inbound connections are
proxied from ports forwarded to the hosts interface on the external network.

\subsection{Implementation}
Based on the above concept we can develop an architecture for an implementation
of a \emph{rosproxy}. The concept is shown in \cref{rosproxy:img}. At runtime,
the proxy will consist of multiple individual TCP proxies and multiple HTTP
endpoints as XMLRPC proxies. A \texttt{/master} endpoint will proxy all calls to
the master API from nodes on the internal network \emph{br}. For each node (e.g.
node \texttt{/caller\_id}) a new HTTP endpoint will be allocated dynamically
(e.g. \texttt{/node/caller\_id}). This node specific endpoint will expose a
nodes slave API via the proxy. For each TCPROS endpoint reported by a node, a
TCP proxy -- forwarding a port on the \emph{rosproxy} host to the respective
endpoint -- will be instantiated. In addition to an HTTP endpoint, an additional
TCP proxy will be allocated for each nodes slave API. All the resources
allocated per individual node are shown enclosed by a gray dashed line in
\cref{rosproxy:img}. Lifetime of those resources is managed by counting a nodes
subscriptions, publications and registered services, as well as cyclic "pinging"
of the node, using its XMLRPC API. This resource lifetime management in combination with
the additional TCP proxy for each nodes slave API enables support for stale node
detection an registration removal, as implemented by \texttt{rosnode
cleanup}\cite{rosnodeROSWiki}.
The internal handler functions for each HTTP endpoint will take care of
allocating new endpoints and TCP proxies, and rewriting those
addresses/endpoints in XMLRPC calls.

While the diagram in \cref{rosproxy:img} focuses on the operating principle and
shows proxying of ROS communication from an internal to an external network, we
can see the core functionality enabling containerization of ROS nodes: The
exemplary ports $p_1$ and $p_2$ can be allocated from a predefined range
($[x,y]$ in \cref{container:img:rosproxy}) and exposed via the host. In the
intended setup as shown in \cref{container:img:rosproxy}, the \emph{rosproxy} in
turn has to use (and can be configured to do so) the hosts hostname or address
for the reported and rewritten XMLRPC and TCPROS endpoints.

We implement \emph{rosproxy} using \emph{JavaScript} (\emph{ECMA Script}),
targeting the \emph{Node.js} runtime. The implementation is available at
\cite{rosproxy}.

\section{Conclusion}
We analyzed the networking-related challenges while containerizing ROS nodes in
distributed multi-host environments for mixed deployment of containerized nodes
and nodes ran directly on a host machines OS. Focusing on \emph{docker} as
container runtime, we provide a solution to the problem of bidirectional
connection establishment, requiring no shared configuration management, no
modification to a stock \emph{docker} installation or changes to the underlying
infrastructure and container host, by enabling the proposed proxy itself to be
ran inside a container. Despite the focus on the \emph{docker} runtime, the
presented solution is runtime agnostic. We believe the work to be of great value
by enabling fast and easily reproducible deployments of ROS systems on
heterogenous infrastructure, by leveraging containerization and container
distribution mechanisms. Simultaneously lowering the maintenance and management
overhead for these systems, as containers can ship all required dependencies.
While this work targets ROS and not ROS 2, we believe that enabling evaluation
and research into containerized ROS deployments will deliver value beyond the
lifetime of ROS and into ROS 2. The ROS 2 DDS/RTSP-based \cite{ROSDDS} transport
and discovery may in the future lend itself to replicating similar setups as
implemented and discussed here, by e.g. using DDS routing services as shown in
\cite{DDSDocker}.

While we set out to only enable containerization of ROS nodes, we effectively
provide a solution to proxy ROS communication over network boundaries and
between network segments in general. With this capability, we imagine the work
to enable further research in cloud robotics, by enabling transparent proxying
for ROS, using its native communication protocols and technologies.





\bibliographystyle{IEEEtran}
\IfFileExists{./bib/bibliography_final.bib}{%
  \typeout{using: bibliography_final.bib}%
  \bibliography{IEEEabrv,./bib/bibliography_final}%
  }{%
    \IfFileExists{./bib/bibliography.bib}{%
        \typeout{using: bibliography.bib}%
        \bibliography{IEEEabrv,./bib/bibliography}%
      }{%
				\typeout{bibliography not found, using empty!}%
				\bibliography{IEEEabrv,./bib/empty}%
			}%
	}%

\end{document}